%% file: main.tex
\documentclass[letterpaper, 10pt, conference]{ieeeconf}      

\IEEEoverridecommandlockouts                              

\usepackage{etoolbox}
\usepackage{capt-of}
\makeatletter
\let\NAT@parse\undefined
\apptocmd\@maketitle{{\eyecatcher{}\par}}{}{}
\makeatother
\usepackage{graphics} 
\usepackage{epsfig} 
\usepackage{times} 
\usepackage{mathtools,amsmath} 
\usepackage{amssymb}  
\usepackage[bookmarks=true]{hyperref}
\usepackage{xcolor,colortbl}
\usepackage{color}
\usepackage{siunitx}
\usepackage{multirow}
\usepackage{multicol}
\usepackage[skip=2pt,font=small]{caption}
\usepackage{subcaption}
\usepackage{booktabs}
\usepackage{pifont}
\usepackage{cite}
\usepackage{bm}
\usepackage{soul}

\usepackage{tikz}
\usetikzlibrary{svg.path,external}
\usepackage[utf8]{inputenc}
\usepackage{pgfplots}
\DeclareUnicodeCharacter{2212}{−}
\usepgfplotslibrary{groupplots,dateplot}
\usetikzlibrary{patterns,shapes.arrows}
\pgfplotsset{compat=newest}

\newcommand\eyecatcher{
\centering
\includegraphics[width=0.8\linewidth]{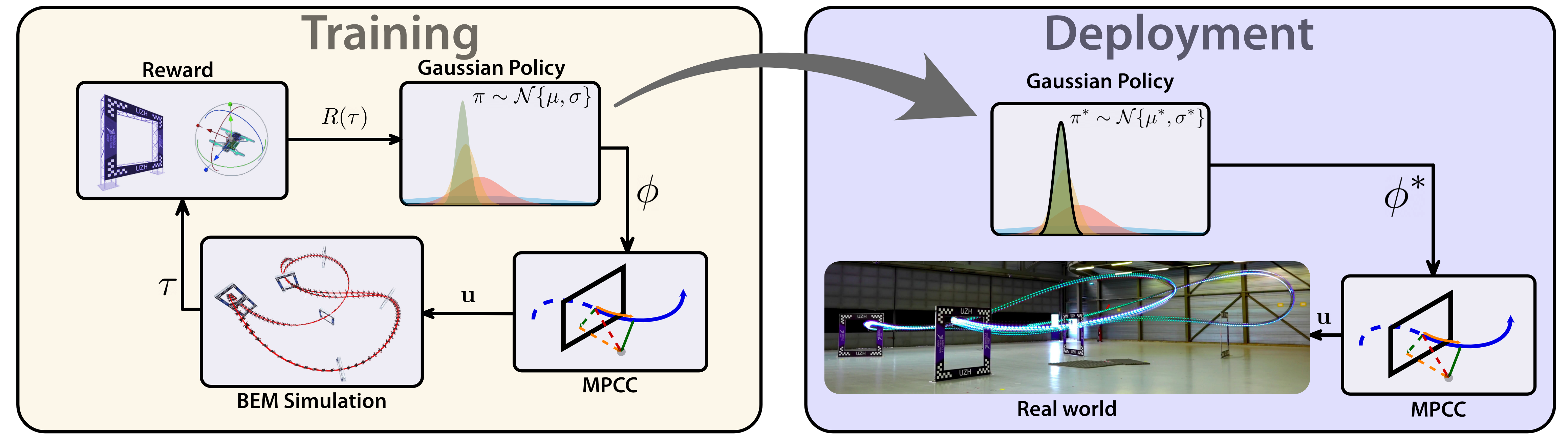}
\captionof{figure}{\rebuttal{During training, we sample a set of tuning parameters $\vec{\phi}$, run a series of simulations with a Model Predictive Contouring Controller in the loop, and obtain a set of trajectories $\tau$. The reward signal computed from $\tau$ is then used to update the policy. The best policy is directly deployed in the real world and validated against both the hand-tuned controller and the state-of-the-art auto-tuning baseline. Our approach outperforms these methods and achieves an unprecedented lap time, reaching speeds of~\SI{75}{\kilo\meter\per\hour}.}}
\label{fig:eyecatcher}
\vspace*{-11pt}
}

\newcommand{\rebuttal}[1]{#1}

\newcommand{\bs}{\boldsymbol}

\newcommand{\hide}[1]{}
\let\vec\bm
\newcommand{\mat}[1]{\begin{bmatrix}#1\end{bmatrix}}

\DeclareMathOperator*{\argmin}{argmin}
\let\vec\bm

\DeclareMathOperator*{\argmax}{\arg\!\max}
\definecolor{Gray}{gray}{0.9}
\definecolor{somegray}{rgb}{0.5, 0.5, 0.5}
\newcommand{\darkgrayed}[1]{\textcolor{somegray}{#1}}

 \makeatletter
 \newcommand*\titleheader[1]{\gdef\@titleheader{#1}}
 \AtBeginDocument{%
   \let\st@red@title\@title
   \def\@title{%
     \vskip-4em
     \bgroup\normalfont\large\centering\@titleheader\par\egroup
     \vskip0.5em\st@red@title}
 }
\makeatother

 \titleheader{ \darkgrayed{This paper has been accepted for publication at the \\
  2023 IEEE International Conference on Robotics and Automation (ICRA)
 \copyright IEEE} }

\title{Weighted Maximum Likelihood for\\
Controller Tuning}
\author{Angel Romero*, Shreedhar Govil*, Gonca Yilmaz*, Yunlong Song, Davide Scaramuzza
    \thanks{
    *Equal contribution. The authors are with the Robotics and Perception Group, Department of Informatics, University of Zurich, and Department of Neuroinformatics, University of Zurich and ETH Zurich, Switzerland (\protect\url{http://rpg.ifi.uzh.ch}). This work was supported by the Swiss National Science Foundation (SNSF) through the National Centre of Competence in Research (NCCR) Robotics, the European Union’s Horizon 2020 Research and Innovation Programme under grant agreement No. 871479 (AERIAL-CORE), and the European Research Council (ERC) under grant agreement No. 864042 (AGILEFLIGHT).
    }
}

\begin{document}

\maketitle
\input{Sections/abstract}
\vspace{-8pt}
\section*{Supplementary Material}
Video of the experiments: \url{https://youtu.be/21junNZ-M38}
\vspace{-12pt}
\input{Sections/introduction}
\input{Sections/relatedwork}
\input{Sections/methodology}
\input{Sections/experiments}
\input{Sections/conclusion}
\bibliographystyle{IEEEtran}
\bibliography{references}
\input{Sections/appendix}
\end{document}

%% file: Sections/abstract.tex
\begin{abstract}
%
%
%
Recently, Model Predictive Contouring Control (MPCC) has arisen as the state-of-the-art approach for model-based agile flight.
%
%
MPCC benefits from great flexibility in trading-off between progress maximization and path following at runtime without relying on globally optimized trajectories. 
However, finding the optimal set of tuning parameters for MPCC is challenging because (i) the full quadrotor dynamics are non-linear, (ii) the cost function is highly non-convex, and (iii) of the high dimensionality of the hyperparameter space.
This paper leverages a probabilistic Policy Search method---Weighted Maximum Likelihood~(WML)---to automatically learn the optimal objective for MPCC.
WML is sample-efficient due to its closed-form solution for updating the learning parameters. 
%
%
Additionally, the data efficiency provided by the use of a model-based approach allows us to directly train in a high-fidelity simulator, which in turn makes our approach able to transfer zero-shot to the real world.
We validate our approach in the real world, where we show that our method outperforms both the previous manually tuned controller and the state-of-the-art auto-tuning baseline reaching speeds  of  75 km/h.



\end{abstract}

%% file: Sections/introduction.tex
\section{Introduction}
Many fields in modern robotics, such as autonomous robot navigation or legged locomotion, become impeded by the complexity of simultaneously handling multiple task objectives while also dealing with environment changes over time. 
For instance, when flying a vision-based quadrotor, it is essential to include a perception-aware objective in the cost function due to the limited field of view of the camera and the underactuated nature of the quadrotor dynamics.
Another example comes from quadrupedal robots, where the robot needs to keep walking forward despite facing highly irregular profiles, deformable terrain, slippery surfaces, and overground obstructions \cite{marcohutter2020}.

Recent controller designs relying on optimization have shown the potential to cope with multiple task objectives and environment changes. 
Falanga et al. \cite{Falanga18iros} proposed a perception-aware model predictive controller that can simultaneously track a trajectory and maximize the visibility of a point of interest in the camera image. 
Similarly, Model Predictive Contouring Control \rebuttal{\cite{lam2010model, lam2012model} applied to quadrotor flight \cite{mpcc_tro, mpcc_replanning}} addresses the time-optimal flight problem by solving the time-allocation and the contouring control problem simultaneously, in real time, manifesting great adaptation against model mismatches and unknown disturbances.
%
%
%

%

However, despite the flexibility provided by numerical optimization, the design of a well-formulated task objective is challenging. 
To achieve real-time control performance, several approximations are introduced in the cost formulation, resulting in time-consuming hand tuning which leads to sub-optimal performance.
Machine learning can be leveraged to select optimal parameters for optimization-based controllers automatically. 

There are different existing approaches for automatic controller tuning.
Specifically, one approach that has shown promising results in the domain of car racing is \cite{BOZeilinger}, which uses Bayesian Optimization to jointly optimize the parameters of the dynamic model and tuning parameters of an MPCC architecture.
Similarly, AutoTune \cite{autotune} tackles the receding horizon controller tuning for agile flight by leveraging statistical learning. 
It has the benefit that it can directly cope with sparse metrics such as lap times or gate collisions by sampling the metric function using Metropolis-Hasting, an efficient version of Monte Carlo sampling that is non-gradient based. 
%
While AutoTune has shown to outperform existing methods, we show that it has difficulties when dealing with controllers where the dimensionality of the parameter space increases.
Both these related works have only dealt with a relatively small dimensionality in the parameter space (less than 10 parameters).

Recently, Policy-Search methods have successfully been used for designing a Model Predictive Controller (MPC) to perform agile flight within very complex environments \cite{yunlong_policy}. 
The authors exploited the advantages of Policy Search to learn complex policies using data from past experience and applied them to achieve optimal-control performance in the task of agilely flying through fast-moving gates.

In this paper, we propose the use of Policy Search---in particular, Weighted Maximum Likelihood (WML)---for searching the space of hyperparameters of a  Model Predictive Contouring Controller (MPCC).
In contrast to end-to-end learning, where the agent needs to learn everything from data, having an optimal controller in the loop leverages the use of a model by the optimizer, which allows a sample efficiency that permits training in a high-fidelity simulator \cite{bauersfeld2021neurobem}.
We show that the controller has the ability to transfer zero-shot to the real world by flying extremely agile maneuvers.

\subsection*{Contributions}

\rebuttal{
The contributions of this paper are: (i) we show that we are able to find the optimal hyperparameters for a state-of-the-art MPCC controller in the task of time-optimal flight, pushing the performance of the controller to the best lap times to date. We show this by outperforming the best tuning that a human expert was able to achieve after trying for several weeks, in both simulation and the real world; and (ii) we demonstrate the ability to transfer the results obtained in simulation to the real world by flying through two different tracks at speeds up to~\SI{75}{\kilo\meter\per\hour}.
}

%% file: Sections/relatedwork.tex
\section{Related Work}

\subsection{Automatic controller tuning}
There exist different methods for automatically tuning controllers\rebuttal{\cite{schperberg2022auto, zanon2020safe, cheng2022difftune}}.
%
%
In line with adaptive control \cite{hanover2021performance}, the classic approach (known as the MIT rule \cite{MITrule}) for controller tuning analytically finds the relationship between a performance metric, e.g. tracking error or trajectory completion, and optimizes the parameters with gradient-based optimization \cite{grimble1984implicit, aastrom1993automatic, mohd2015intelligent}.
However, expressing the long term measure (in our case the laptime and gate passing metrics) as a function of the tuning parameters is impractical and generally intractable.
Instead of analytically computing it, another line of work proposes to iteratively estimate the optimization function, and use the estimate to find optimal parameters \cite{menner2020maximum, berkenkamp2016safe, marco2016automatic}.
However, these methods make over-simplifying assumptions on the objective function, e.g. convexity or relative Gaussianity between observations. Such assumptions are generally not suited for controller tuning to high-speed flight, where the function is highly non-convex.
%


Other approaches are mainly data-driven and directly focus on using a receding horizon optimal control approach in the center of a learning algorithm.
%
Learning-based MPC~\rebuttal{\cite{edwards2021automatic, kabzan2019learning, spielberg2021neural, chee2022knode, saviolo2022physics, williams2018information, williams2017infoMPC, ostafew2016robust, rosolia2019learning, torrente2021data, mehndiratta2018automated}} can leverage real-world data to improve dynamic modeling or learn a cost function for MPC. 
It allows for a more robust and flexible MPC design.
In particular, sampling-based MPC~\cite{williams2018information} algorithms
are developed for handling complex cost criteria and general nonlinear dynamics.
This is achieved by combining neural networks for the system dynamics approximation with
the model predictive path integral (MPPI) control framework~\cite{williams2018information} for real-time control optimization. 
A crucial requirement for the sampling-based MPC is to generate a large number of samples 
in real-time, where the sampling procedure is generally performed in parallel 
by using graphics processing units~(GPUs).
Hence, it is computationally and memory expensive to run sampling-based MPC on embedded systems.
These methods generally focus on learning dynamics for tasks where a dynamical model of the robots
or its environment is challenging to derive analytically, such as 
aggressive autonomous driving around a dirt track~\cite{williams2017infoMPC}.
More recent approaches\rebuttal{\cite{amos2018differentiable, cheng2022difftune, theseus}} pose the optimization problem as an additional differentiable layer through which one can \rebuttal{apply gradient-based optimization methods, e.g. back-propagation}.
In \cite{amos2018differentiable}, they are able to directly find the gradient of the MPC controller with respect to its high level hyperparameters. They use the KKT (Karush-Kuhn-Tucker) conditions of the convex approximation to find this gradient.
This allows treating MPC controllers as an extra layer in an end-to-end learning structure.
\rebuttal{Another work in this regard applies reverse mode auto-differentiation and proves to be robust to uncertainties in the system dynamics
and environment in simulation\cite{cheng2022difftune}}.

\subsection{High-speed quadrotor flight}
The literature on high-speed, high agile quadrotor starts with polynomial planning and classical control. 
The main focus was directed to exploiting the differential flatness property of the quadrotor and leverage the use of polynomial for planning \cite{Mueller11iros, Mahony12ram, Mellinger12ijrr, Mueller13iros}.

More recently, optimization-based methods have achieved true time-optimal trajectory planning using the quadrotor dynamics and numerical optimization~\cite{foehn2021CPC}. 
However, due to high computational complexity, it normally requires several minutes or hours to solve, and the trajectory can only be computed offline.
Additionally, a small deviation from the planned trajectory leads to catastrophic crashes since the vehicle is already operating on the edge of its actuator limits. 
The solution to this problem is the current state-of-the-art minimum-time flight: Model Predictive Contouring Control~(MPCC)~\cite{mpcc_tro}, which simultaneously does online time-optimal trajectory planning and vehicle control.
In contrast to standard trajectory tracking control (e.g., MPC), MPCC balances the maximization of the progress along a given path and minimizes the deviation from it.
MPCC has the freedom to optimally select at runtime the states at which the reference trajectory is sampled such that the progress is maximized while the platform stays within actuator bounds.

Latest developments in quadrotor simulations have allowed for evaluation environments where we can not only train but also evaluate and zero-shot transfer control policies to the real-world.
The state of the art on realistic simulations for quadrotors is \cite{bauersfeld2021neurobem}, which introduces a hybrid aerodynamic quadrotor model which combines blade element momentum theory with learned aerodynamic representations from highly aggressive maneuvers.

%% file: Sections/methodology.tex
\section{Preliminaries}
In this section, we introduce the basics of \rebuttal{Policy Search followed by} Model Predictive Contouring Control. Then, we briefly introduce quadrotor dynamics.
\vspace{-6pt}
\rebuttal{
\subsection{Policy Search}
\label{sec:policy_search}
We summarize policy search by following the derivation from~\cite{yunlong_policy}. In particular, we focus on episode-based policy search (or episodic reinforcement learning).
%
%
Episode-based policy search adds perturbations in the policy parameter space instead of the policy output.
A reward function~$R(\bm{\tau})$ is used to evaluate the quality of trajectories~$\bm{\tau}$ that are generated by sampled parameters~$\bm{\phi}$.
In our case, we search for a constant Gaussian policy  where the set of MPCC parameters $\phi$ that maximizes the reward is found.
%
Therefore, the policy parameters~$\bm{\phi}$ are updated by maximizing the 
expected return of sampled trajectories
\begin{equation}
    J_{\bm{\phi}} =\mathbb{E}[R(\bm{\tau}) | \bm{\phi}] \approx \int R(\bm{\tau}) p_{\bm{\phi}}(\bm{\tau}) d\bm{\tau}.
    \label{eq:general_expectation}
\end{equation}
%
%
%
%
We focus on a probabilistic model in which the search for high-level decision variables $\vec{\phi}$ in the MPCC optimization is treated as a probabilistic inference problem, and use the Expectation-Maximization algorithm to update the policy parameters.
%
%
We use a Gaussian distribution~$\mathbf{z} \sim\pi_{\bm{\phi}}=\mathcal{N}(\bm{\mu}, \bm{\Sigma})$ to represent the policy, where $\bm{\mu}$ is a mean vector and  $\bm{\Sigma}$ is a diagonal covariance matrix. 
The covariance matrix is needed in order to incorporate exploration. 
Therefore, the MPCC parameters are~$\bm{\phi}=[\bm{\mu, \Sigma}]$.    
}

\subsection{Model Predictive Contouring Control}\label{sec:mpcc}
\rebuttal{Consider the discrete-time dynamic system with continuous state and input spaces, $\vec{x}_k \in \mathcal{X}$ and $\vec{u}_k \in \mathcal{U}$ respectively. 
Let us denote the time discretized evolution of the system \mbox{$f : \mathcal{X} \times \mathcal{U} \mapsto \mathcal{X}$} such that $\vec{x}_{k + 1} = f(\vec{x}_k, \vec{u}_k)$.}

%
%
In contrast to traditional MPC approaches, where the objective is to track a dynamically feasible trajectory, the main objective of MPCC is to maximize the predicted traveled distance along a path.
This idea is encoded in the controller by introducing the concept of \emph{progress}, which is the travelled arc-length measured along the nominal path.
In this paper, we aim to tune this MPCC controller for the task of drone racing. To that end, we consider the MPCC problem in its full form:
\begin{align}
  \pi(\vec{x}) = & \argmin_u
  & & \begin{multlined}\sum_{k = 0}^{N } \Vert \vec{e}^l(\theta_k) \Vert_{q_l}^2 + \Vert \vec{e}^c(\theta_k) \Vert_{q_c}^2 + \Vert\vec{\omega}_k\Vert_{\bm{Q_{\omega}}}^2 \\ + \Vert\Delta v_{{\theta}_k}\Vert_{r_{\Delta v}}^2 + \Vert\Delta\vec{f}_k\Vert_{\bm{R_{\Delta f}}}^2 - \mu v_{\theta,k}
  \end{multlined} \notag \\
  & \text{subject to} & & \vec{x}_0 = \vec{x} \notag \\
  &&&\vec{x}_{k+1} = f(\vec{x_k}, \vec{u_k}) \notag \\
  &&&\vec{g_{lb}} \leq \vec{g(x_k, u_k)} \leq \vec{g_{ub}}
  \label{eq:full_ocp}
\end{align}
%
where $\vec{g_{lb}}$ and $\vec{g_{ub}}$ encode the lower and upper bounds on states and inputs, $\theta_N$ is the progress at the last horizon step, and \rebuttal{$e^c_k$ is the contour error at time $k$, which is the position error between the current position and the nominal path. For more details about the problem definition we refer the reader to \cite{mpcc_tro}.}

%


\begin{figure}[th!]
\centering
\includegraphics[width=0.6\columnwidth]{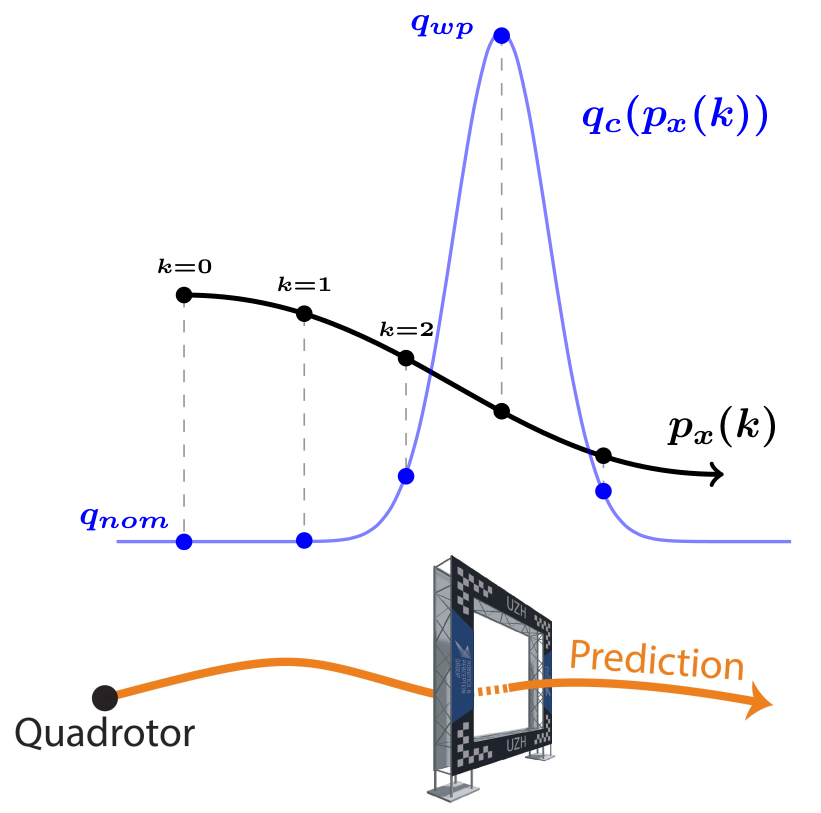}
\caption{Dynamic allocation of the weight in the $x$ axis. When closer to the gate, the contouring weight $q_c(p_x(k))$ (shown in blue) grows to $q_{wp}$, decreasing the tracking error around the gate. Everywhere else $q_c$ is kept constant and equal to $q_{nom}$ such that the MPCC has more freedom to optimize for progress.
}
\label{fig:gaussians}
\vspace*{-8pt}
\end{figure}
%
%
This MPCC algorithm has been adapted for the task of drone racing by adding a varying state-dependent cost to the $q_c$ parameters, in the form of a scaled Gaussian distribution located at every \rebuttal{waypoint (or gate)} that needs to be passed. For every gate, the height $h$ and the width $w$ of the Gaussian are defined, which are additional tuning parameters. This way, a region of attraction at every gate is created. Therefore, the set of tunable parameters $\phi$ are:
\begin{align*}
    \vec{\phi} = 
    \begin{bmatrix}
    h_0, w_0, \dots, h_{n_g-1}, w_{n_g-1}, q_{nom}, r_{\Delta v}, r_{\Delta f}, \mu
    \end{bmatrix}
\end{align*}
where $n_g$ is the total number of gates. The dimensionality of the hyperparameter space is therefore $2n_g + 4$. The cost function of optimization problem \eqref{eq:full_ocp} can then be written as $J_{\vec{\phi}}(\vec{x}, \vec{u})$, where we need to find the set of optimal hyperparameters $\vec{\phi^*}$ that solve a given higher level task, which will be encoded by a reward function $R(\vec{\tau})$.
\vspace*{-8pt}
\subsection{Quadrotor Dynamics}
\label{sec:quad_dynamics}
In this section, we describe the specific dynamics used, referred in \eqref{eq:full_ocp} as $f(\vec{x_k}, \vec{u_k})$.
The quadrotor's state space is described from the inertial frame $I$ to the body frame $B$, as $\vec{x} = [\vec{p}_{IB}, \vec{q}_{IB}, \vec{v}_{IB}, \vec{w}_{B}]^T$ where $\vec{p}_{IB}\in \mathbb{R}^3$ is the position, $\vec{q}_{IB} \in \mathbb{SO}(3)$ is the unit quaternion that describes the rotation of the platform, $\vec{v}_{IB} \in \mathbb{R}^3$ is the linear velocity vector, and $\vec{\omega}_{B} \in \mathbb{R}^3$ are the bodyrates in the body frame.
The input of the system is given as the collective thrust $\vec{f}_B = [0~~0 ~~f_{Bz}]^T$ and body torques $\vec{\tau}_B$.
For readability, we drop the frame indices as they are consistent throughout the description.
The dynamic equations are
\begin{gather}
\begin{aligned}
\dot{\vec{p}} &= \vec{v} & \dot{\vec{v}} &= \vec{g} + \frac{\mathbf{R}(\vec{q}) \vec{f}_T}{m} - \bm{R}(\bm{q}) \bm{D} \bm{R}^\intercal(\bm{q}) \cdot \bm{v}\\
\dot{\vec{q}} &= \frac{\vec{q}}{2} \odot [0~~\vec{\omega}]^T &
\dot{\vec{\omega}} &= \mathbf{J}^{-1} \left( \vec{\tau} - \vec{\omega} \times \mathbf{J} \vec{\omega} \right)
\label{eq:quad_dynamics}
\end{aligned}
\end{gather}
where $\odot$ represents the Hamilton quaternion multiplication, $\mathbf{R}(\vec{q})$ the quaternion rotation, $m$ the quadrotor's mass, and $\mathbf{J}$ the quadrotor's inertia.

Additionally, the input space given by $\vec{f}$ and $\vec{\tau}$ is decomposed into single rotor thrusts $\vec{f} = [f_1, f_2, f_3, f_4]$ where $f_i$ is the thrust at rotor $i \in \{ 1, 2, 3, 4 \}$.
\begin{align}
\vec{f}_T &= \mat{0 \\ 0 \\ \sum f_i} &
\text{and }
\vec{\tau} &=
\mat{l/\sqrt{2} (f_1 + f_2 - f_3 - f_4) \\
l/\sqrt{2} (- f_1 + f_2 + f_3 - f_4) \\
c_\tau (f_1 - f_2 + f_3 - f_4)}
\label{eq:quad_inputs}
\end{align}
with the quadrotor's arm length $l$ and the rotor's torque constant $c_\tau$.
In order to approximate the most prominent aerodynamic effects, the quadrotor's dynamics include a linear drag model \cite{Faessler18ral}. In \eqref{eq:quad_dynamics}, $\bm{D}$ is the diagonal matrix with drag coefficients $d$ such that $\bm{D} = \text{diag}\left(d_x, d_y, d_z\right)$.
\section{Problem Formulation}
\label{sec:problem_formulation}
\rebuttal{In this section, we combine the approaches introduced in Sections \ref{sec:policy_search} and \ref{sec:mpcc}.}
We treat MPCC as a controller~$\bm{\tau} =\text{MPCC}(\mathbf{z})$ that is parameterized by the high-level decision variables~$\bm{z}$. 
Here, $\bm{\tau} = [\mathbf{u}_h, \mathbf{x}_h]_{h\in{1, \cdots, H}}$ is a trajectory generated by MPCC given~$\mathbf{z}$, where
$\mathbf{u}_h$ are control commands and $\mathbf{x}_h$ are corresponding states of the robot. 
By perturbing~$\mathbf{z}$, MPCC can result in completely different trajectories $\bm{\tau}$.
\rebuttal{For the MPCC to execute the given task in minimum time}, the optimal $\mathbf{z}$ has to be defined in advance.
%

To formulate the policy search as a latent variable inference problem, similar to~\cite{yunlong_policy}, we introduce a binary \emph{reward event} as an observed variable, denoted as $E = 1$.
Maximizing the reward signal implies maximizing the probability of this \emph{reward event}.
The probability of this reward event is given by~$p(E | \bm{\tau}) \propto \exp{ \{ R(\bm{\tau}) \} }$,
where $R(\bm{\tau})$ is a reward function for evaluating
the goodness of the MPCC solution $\bs{\tau}$ with respect to a given evaluation metric of the task.
This leads to the following maximum likelihood problem~\cite{deisenroth2013survey}:
\begin{equation}\label{eq: max_log_pro}
    \max_{\bs{\phi}} \quad \log p_{\bs{\phi}}(E=1) = \log \int_{\bs{\tau}} p(E|\bs{\tau}) p_{\bs{\phi}} (\bs{\tau)} d \bs{\tau},
\end{equation}
which is intractable to solve directly and can be approximated efficiently using Monte-Carlo Expectation-Maximization~(MC-EM)~\cite{kober2009policy, vlassis2009model}. 
MC-EM algorithms find the maximum likelihood solution for
the log marginal-likelihood~(\ref{eq: max_log_pro}) by introducing a variational distribution~$q(\bs{\tau})$, and then, decompose the marginal log-likelihood into two terms:
\begin{equation}
    \log p_{\bs{\theta}}(E=1) = \mathcal{L}_{\bs{\theta}} (q(\bs{\tau}) ) + D_\text{KL}(q(\bs{\tau}) || p_{\bs{\theta}}(\bs{\tau} | E) )
\end{equation}
where the~$D_\text{KL}$ is the Kullback–Leibler~(KL) divergence 
between $q(\bm{\tau})$ and the reward-weighted trajectory distribution~$p_{\bs{\theta}}(\bs{\tau} | E)$. 
%

The MC-EM algorithm is an iterative method that alternates between 
performing an Expectation~(E) step and a Maximization~(M) step.
In the expectation step, we minimize the KL-divergence~$D_\text{KL}$,
which is equivalent to setting $q(\bs{\tau})=p_{\bs{\theta}}(\bs{\tau} | E) \propto p(E|\bs{\tau})p_{\bs{\theta}}(\bs{\tau})$.
In the maximization step, we update the policy parameters by maximizing the 
expected complete data log-likelihood
\begin{equation}
    \bs{\phi}^{\ast} = \arg \max_{\bs{\phi}} \sum_i p(E | \bs{\tau}^{[i]}) \log p_{\bs{\phi}} (\bs{\tau}^{[i]})
\end{equation}
where each sample $\bs{\tau}^{[i]}$ is weighted by the probability of the \emph{reward event}, denoted as $p(E|\bs{\tau})$.
The trajectory distribution $p_{\bs{\theta}} (\bs{\tau}^{[i]})$ can be replayed by the high-level policy~$\pi_{\bm{\theta}}$.
To transform the reward signal $R(\bs{\tau}^{[i]})$ of a sampled trajectory $\bs{\tau}^{[i]}$ into a probability distribution of the \emph{reward event}, we use the exponential  transformation~\cite{neumann2011variational, toussaint2009robot, deisenroth2013survey}:
\begin{equation}
    \label{eq: d}
    d^{[i]} =p(E|\bs{\tau}) = \exp{ \left\{ \beta R(\bs{\tau}^{[i]}) \right\} }
\end{equation}
where the parameter $\beta \in \mathbb{R}_{+}$ denotes the inverse temperature of the soft-max distribution, higher value of $\beta$ implies a more greedy policy update. 
\begin{figure}[t!]
    \centering
    \includegraphics[width=0.9\columnwidth]{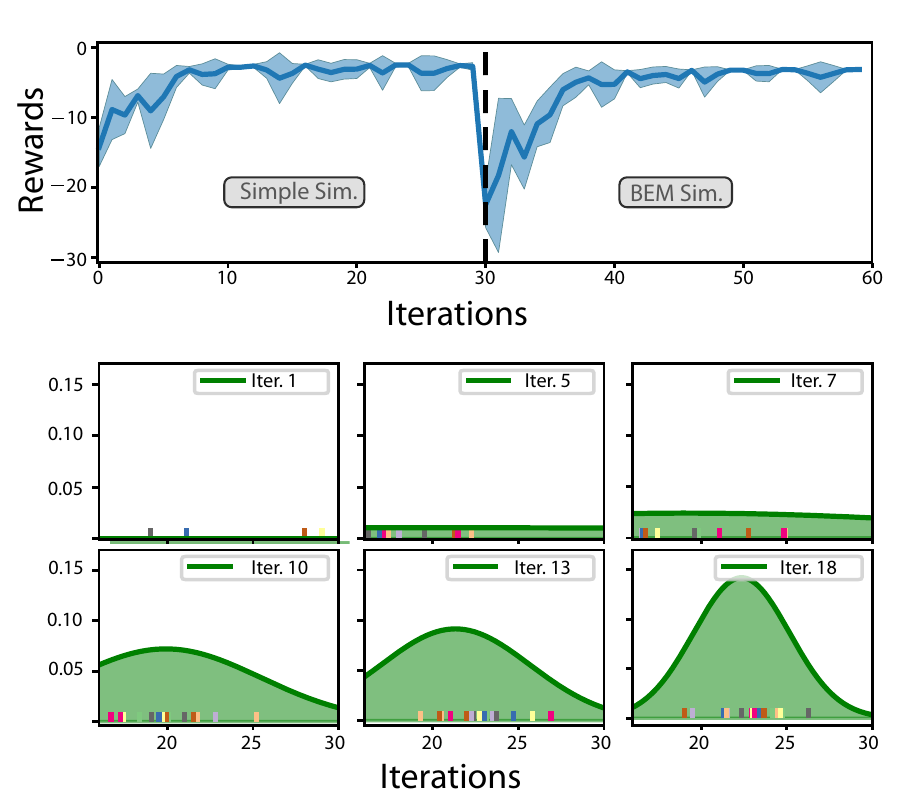}
    \caption{Top: reward evolution as training progresses. The rewards have been computed for 5 training runs averaged. In the middle of the training we switch from simple simulator to BEM simulator. This way, we leverage the faster running time in simple simulation and transfer the policy from simple to BEM. Bottom: example of how a the policy evolves and converges for a single parameter through the iterations.}
    \label{fig:iterations}
    \vspace*{-16pt}
\end{figure}
\begin{table*}[t!]
\footnotesize
\centering
\caption{We show the results obtained in simulation in both the Split-S track and in the Marv track. Our approach is able to find optimal solutions in both tracks, whereas the baselines struggle to find solutions in high dimensional hyperparameter spaces.} \label{tab:results_overview}
\begin{tabular}{l|cc|cc|cc|cc|cc} 
\cmidrule[\heavyrulewidth]{2-9}
& \multicolumn{4}{c|}{\textbf{Split-S track} (18 hyperparameters)}
& \multicolumn{4}{c|}{\textbf{Marv track} (30 hyperparameters)}\\
\cmidrule{2-9}
    &\multicolumn{2}{c|}{ Lap Time $[$s$]$}
    &\multicolumn{2}{c|}{ Success Rate $[$\%$]$}
    &\multicolumn{2}{c|}{ Lap Time $[$s$]$}
    &\multicolumn{2}{c|}{ Success Rate $[$\%$]$}\\
    {} & {Simple} & {BEM} & {Simple} & {BEM} & {Simple} & {BEM} & {Simple} & {BEM} \\ \midrule
    Random  & 6.024 ± 0.519 & 6.031 ± 0.517 & 34 & 27 
    & 10.45 & 10.48 & 1 & 1 \\
    Hand-Tuned & 5.24  & 5.42 & -  & -   
    & 9.17  & Collision & -  & - \\
    AutoTune  & 6.111 ± 0.342  & 6.471 ± 0.473 & 63 & 50
    & Collision  & 9.08 & 0 & 1 \\
    \textbf{WML (Ours)}  & \textbf{5.02 ± 0.008} & \textbf{5.274 ± 0.012} & \textbf{99}  & \textbf{90} 
    & \textbf{8.339 ± 0.106} & \textbf{8.381 ± 0.022} & \textbf{100}  &\textbf{100} \\ \bottomrule
\end{tabular}
\vspace*{-8pt}
\end{table*}
\subsection{Learning-Based Hyperparameter Tuning for MPCC}
As explained in Section \ref{sec:problem_formulation}, to find the optimal hyperparameters $\vec{\phi^*}$, we need to sample a set of trajectories $\vec{\tau}^{[i]}$ for each episode.
We start with a random guess for $\pi_{\vec{\phi}}$, then we sample $N$ different $\phi^{[i]}$. 
From every $\phi^{[i]}$, we run a simulation experiment that consists of completing $L$ laps with the MPCC in the loop and a reward $R(\tau^{[i]})$ is calculated. After all trajectories $\tau^{[i]}$ are completed an update step can be made.
%
%

We designed a reward function specifically for the drone racing task such that it results in robust learning and can be directly used to find the optimal policy in the large action space for different tracks. 
\rebuttal{The reward function choice is driven by three objectives: minimizing the lap time, avoiding gate collisions, and maximizing the number of gates passed.}
We define the reward function as follows.
\begin{gather}
\begin{aligned}
\pi(\bm{\phi}) &= \argmax_{\bm{\phi}} \mathbb{E} \left[ \sum_{i=0}^{N} R(\vec{\tau}^{[i]}) \right]
\end{aligned}
\end{gather}
where $R(\vec{\tau}^{[i]})$ is defined as:
\begin{equation}
    R\left(\vec{\tau}^{[i]}\right)
    =
    - t_{1} - 2t_{2} - 2^{{10}^{r_{miss}}} + r_{pass} +  \frac{n_{gp}}{n_{g}} 
    \label{eq:reward_function}
\end{equation}
where $t_1$ and $t_2$ are the lap times of the first and second lap respectively, $n_{gp}$ is the number of gates passed, $n_g$ is the total number of gates, and $r_{miss}$ is the sum of gate passing errors and 
\begin{align}
    &\quad r_{pass} =           \begin{cases}
        1 & \text{if } r_{miss} \leq 0.01 \\
        0 & \text{otherwise } \\
      \end{cases}
      \label{eq:r_pass}
\end{align}
%
The miss reward, $r_{miss}$, is added exponentially in order to heavily penalize if the quadrotor passes the gate outside of a given threshold, which is in our case $0.5$ m.
We do not add  $r_{miss}$ as a linear component to ensure that the optimal solution does not miss the gates and can optimize for lower lap time, since this error can be very small compared to other terms in the reward \eqref{eq:reward_function}.
During the training $r_{miss}$ can never be greater than 1.5 times the gate passing threshold hence the exponential never explodes. 
%
%
If the quadrotor passes the gate with a distance larger than 1.5 times the gate passing threshold, then we count this as a gate crash and give a large negative reward. If it passes all the gates with zero error we give a positive reward in $r_{pass}$ \eqref{eq:r_pass}.
Additionally, we give higher weight to the $t_2$ than $t_1$ because the first lap is slower than the second one and the training prefers to decrease $t_1$ more. 
By giving a higher weight to the $t_2$ we decrease them both simultaneously.



%% file: Sections/experiments.tex
\section{Experiments}
\subsection{Baselines}
For the baselines, we choose three different approaches.
First, we compare against the complete random approach, where we randomly sample the parameters.
Second, we compare against the best results achieved by the previous literature where the controller was tuned by hand by an expert \cite{mpcc_tro}. 
And third, we implement AutoTune, the current state-of-the-art in controller tuning \cite{autotune} for our MPCC controller.
All approaches have the same initialization strategy and allowed search space.
Both AutoTune and our approach have exactly the same reward function, which is described in Eq. \eqref{eq:reward_function}.
\begin{figure*}[t!]
    \centering
    \includegraphics[width=0.75\textwidth]{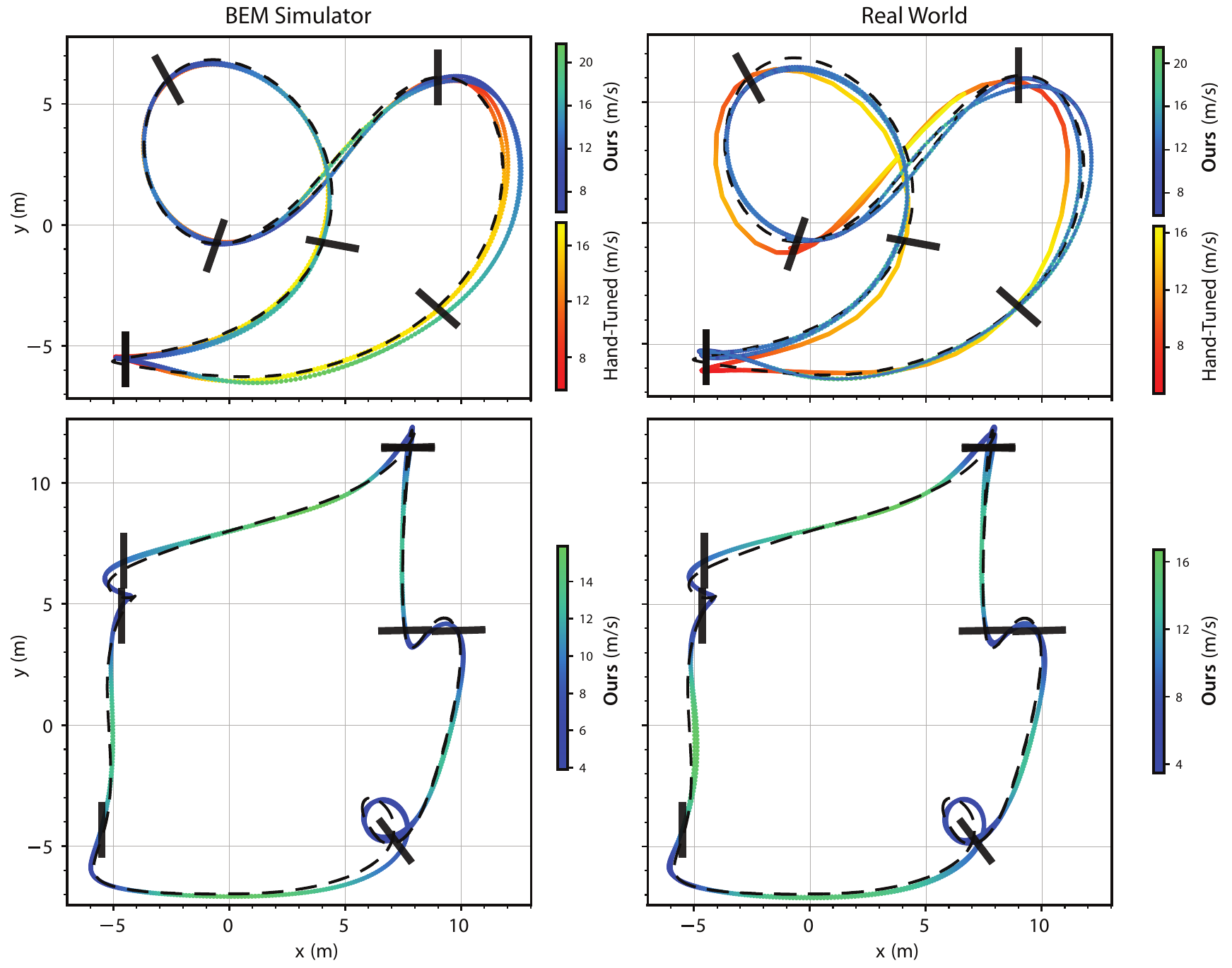}
    \caption{Top: Split-S track. Bottom: Marv track. References are shown in dashed. Comparison of hand-tuned and our method, in simulation and in the real-world. The speeds achieved by our method are consistently higher than those achieved by the hand-tuned approaches. In fact, the human expert was not able to hand-tune a controller that was able to fly the Marv track. The tuning in simulation transfers to the real-world in both tracks.}
    \label{fig:sim_real_world_results}
    \vspace*{-18pt}
\end{figure*}
\subsection{Training}
The training experiments are conducted using \rebuttal{\emph{Flightmare} \cite{song2020flightmare}} and two different simulation environments \rebuttal{for each track}. 
One called \emph{Simple} simulator, where the drone model is the one described in Section \ref{sec:quad_dynamics} and does not capture complex aerodynamic environments. 
The other one is a high-fidelity simulator called \emph{BEM} \cite{bauersfeld2021neurobem}, which includes a high-complexity aerodynamic simulation derived from blade element momentum theory.
%
%

During training, for each sample, we simulate $L = 2$ consecutive laps. We check whether the quadcopter is flying through all the gates and measure the lap times in order to compute the reward \eqref{eq:reward_function}.
We sample $N = 16$ set of parameters from the same Gaussian policy, and then do a policy update step for the episode.
%
During training, our criterion of gate passing is that the drone passes within $0.5 m$ of the gate center.

%
%
When we train directly in the BEM simulator from a random initialization, it usually takes longer to converge or it does not necessarily find the optimal solution.
To address this, we first train in the simple simulator for 30 episodes and then initialize the training in BEM from the solution obtained in the Simple simulator. 
We reset the variance when we begin the training in the BEM environment so that the policy recovers its full exploration capabilities and perform another 30 episodes in BEM. Each episode takes around 35 seconds in Simple simulator and 200 seconds in BEM simulator around. %
The evolution of the rewards in Simple and BEM is shown in Figure \ref{fig:iterations}.

\subsection{Results in Simulation}
In this section, we present the results achieved by our approach when compared against the baselines for two different tracks.
The reference for the Split-S track has been generated with \cite{foehn2021CPC} while for the Marv track, we used \cite{song2021autonomous}.
For the comparison, we run 10 trials with different random initializations with the three approaches (Random, AutoTune and Ours).
We draw 100 samples from our policy and the best 100 samples from Random and AutoTune. We present the results in Table ~\ref{tab:results_overview} and in Fig. ~\ref{fig:sim_real_world_results}. 
We validate our results with gate passing threshold $0.6$ m to the gate center, as the gate width is $1.2$ m in real-world. 
As shown in Table \ref{tab:results_overview}, our approach greatly outperforms both in laptimes, speed, success rate and consistency to all other baselines.
Even for the Marv track, where hand-tuning was impossible due to the complexity of the track and the higher amount of tuning parameters, i.e. 30, our approach finds a set of optimal hyperparameters that fly through the track with 100\% success rate, whereas the baselines struggle to find a successful solution.
%


\subsection{Results in Real World}

We deploy the algorithm to the physical platform, which was built in-house from off-the-shelf components.
A more detailed description of our platform can be found in \cite{agilicious}.

The flying results in real-world for both tracks are shown in Fig. \ref{fig:sim_real_world_results}.
One remark is that the learning-based policy transfers zero-shot to the real-world for both tracks without issue.
We can see this by comparing them in XY position and in maximum velocity (blue colorbars).
There is no hand-tuned data for the Marv track because it was not possible for a human expert to tune a controller that could complete the track without any collision.

Additionally, it is noteworthy to mention that with learning-based approach, we are able to get the fastest lap times ever achieved with the MPCC controller in these tracks. 
For the Split-S track, the fastest lap time of the hand-tuned controller in the real world, as reported in \cite{mpcc_tro}, is of $5.8s$, with a maximum speed of $60$ km/h, whereas the fastest lap time achieved by our learning-based tuning method is $5.29$ s, with a maximum speed of $75$ km/h.
This improvement of around half a second shows how the proposed method can bring out the full potential of the MPCC controller.

%% file: Sections/conclusion.tex
\section{Conclusion}
In this paper we show that by using the proposed learning-based algorithm to tune a model-based controller (in our case a MPCC), we can achieve performance limits that were impossible to achieve using any other approach.
The data efficiency acquired by having a model-based controller in the loop allows us to train directly in a high fidelity simulator, which in turn has shown to allow zero-shot transfer to the real-world.
A possible extension of this work would be to, instead of fixing the regions of attraction to Gaussians (as depicted in Fig. \ref{fig:gaussians}), one could let the data-driven algorithm to directly learn the best appropriate set of weights in the horizon at every time-step.

One limitation of our method is that when starting from a random initialization for $\phi$, in some cases the algorithm leads to non-increasing rewards.
A solution for this is to use a warm-starting strategy. \rebuttal{We partially address this by initiating the training in the simple simulator, then continuing in the BEM simulator}.
\rebuttal{
Another limitation is that we need to re-train for every new track, since the size of the parameter space depends on the number of gates.
This can be tackled by resorting to more complex network architectures and making the parameter set track independent.
}
\rebuttal{Lastly, the reward function is specifically tailored for drone racing objectives and to adapt it to other domains, one would need to re-design it.}
These results open the door to finding new real-world applications for controllers where the high-level objective can be expressed in terms of a reward function directly.
%
